\documentclass[sigconf, nonacm=true]{acmart}
\usepackage{algorithm}
\usepackage{algorithmic}
\usepackage{amsmath}
\usepackage{multirow}
\DeclareMathOperator*{\argmax}{arg\,max}

\usepackage{thmtools}
\usepackage{thm-restate}

\theoremstyle{definition}
\newtheorem{definition}{Definition}

\theoremstyle{remark}
\newtheorem*{remark}{Remark}     
\usepackage{bm}
\usepackage{bbm}
\AtBeginDocument{%
  }

\begin{document}

\title{Auto-bidding in real-time auctions via Oracle Imitation Learning}

\author{Alberto Silvio Chiappa}
\email{alberto.chiappa@epfl.ch}
\affiliation{%
  \institution{Sony*\thanks{*Work done during internship at Sony}, EPFL}
  \country{Japan, Switzerland}
}

\author{Briti Gangopadhyay}
\email{briti.gangopadhyay@sony.com}
\affiliation{%
  \institution{Sony}
  \country{Japan}
  }

\author{Zhao Wang}
\email{zhao.wang@sony.com}
\affiliation{%
  \institution{Sony}
  \country{Japan}
}
\author{Shingo Takamatsu}
\email{shingo.takamatsu@sony.com}
\affiliation{%
  \institution{Sony}
  \country{Japan}
}





\renewcommand{\shortauthors}{Chiappa et al.}

\begin{abstract}
    Online advertising has become one of the most successful business models of the internet era. Impression opportunities are typically allocated through real-time auctions, where advertisers bid to secure advertisement slots. Deciding the best bid for an impression opportunity
    is challenging, due to the stochastic nature of user behavior and the variability of advertisement traffic over time.
    In this work, we propose a framework for training auto-bidding agents in multi-slot second-price auctions to maximize acquisitions (e.g., clicks, conversions) while adhering to budget and cost-per-acquisition (CPA) constraints. We exploit the insight that, after an advertisement campaign concludes, determining the optimal bids for each impression opportunity can be framed as a multiple-choice knapsack problem (MCKP) with a nonlinear objective. We propose an "oracle" algorithm that identifies a near-optimal combination of impression opportunities and advertisement slots, considering both past and future advertisement traffic data.
    This oracle solution serves as a training target for a student network which bids having access only to real-time information, a method we term Oracle Imitation Learning (OIL).
    Through numerical experiments, we demonstrate that OIL achieves superior performance compared to both online and offline reinforcement learning algorithms, offering improved sample efficiency. Notably, OIL shifts the complexity of training auto-bidding agents from crafting sophisticated learning algorithms to solving a nonlinear constrained optimization problem efficiently.
\end{abstract}

\maketitle

\section{Introduction}
\begin{figure}[t]
    \centering
    \includegraphics[width=1\linewidth]{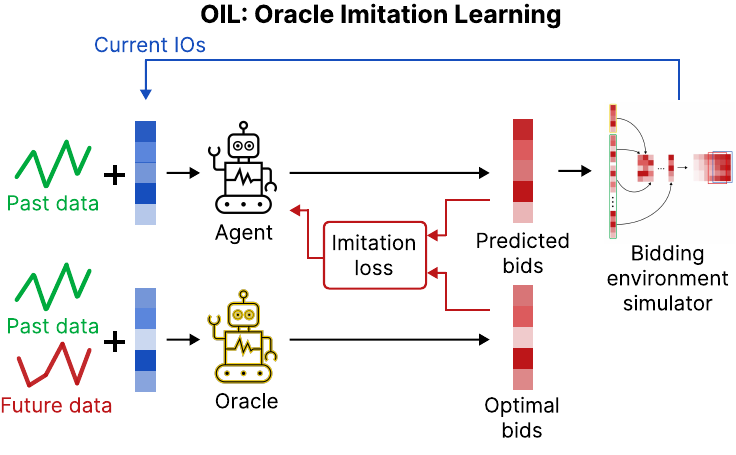}
    \caption{Overview of OIL. At every time step, the auto-biding agent and the oracle observe the conversion probabilities of the available IOs, and bid for each of them. While the agent decides its bids only based on current and past data, the oracle also uses future conversion probabilities. The agent's bids are used to advance the campaign simulation, while the oracle's bids serve as a training target.}
    \label{fig:overview}
\end{figure}

Online advertising is a cornerstone of the internet economy. In the US alone, internet advertising revenues reached \$225 billion in 2023, with a 7.3\% year-over-year increase\footnote{https://www.iab.com/insights/internet-advertising-revenue-report-2024/}. Advertisement impression opportunities (IOs) arise from user actions, such as visiting webpages or scrolling social media feeds~\citep{wang2017display}. Advertisement slots are usually allocated via real-time bidding (RTB)~\citep{muthukrishnan2009ad}, where multiple advertisers compete in an auction to win IOs (e.g., sponsored links in an online search) and show their advertisement~\cite{yuan2013real}. 
Each advertiser can employ a customized bidding strategy, to optimize their specific key performance indicators (KPI), such as engagement, visualizations, clicks and conversions. Bidding strategies adhere to budget and to cost-per-acquisition (CPA) constraints (i.e., how much the advertiser is ready to spend in total and per unit of KPI).  Crafting effective bidding strategies is challenging, due to the stochastic nature of user behavior and the variability of advertisement traffic over time. In this work, we address the problem of maximizing the number of acquisitions (i.e., a generic KPI) in multi-slot bidding, where, for each IO, up to $D$ advertisers can show their advertisement, and each advertiser can win at most one slot. 

Existing methods, including reinforcement learning (RL)~\citep{sutton2018reinforcement} and optimization techniques such as linear programming (LP)~\citep{dantzig2002linear}, have produced effective auto-bidding agents~\citep{,aggarwal2019autobidding,wang2017ladder,zhao2018deep,lu2019reinforcement,he2021unified}. However, they often neglect the global information which becomes available at the end of the advertisement campaign, and could be used to guide better decisions. To this end, we argue that computing which would have been \emph{in hindsight} the optimal bids at each decision step can provide an excellent training target for an auto-bidding agent. In practice, we propose employing an \emph{oracle} agent with access to privileged information (i.e., the advertisement traffic for the whole campaign) to find which are the globally optimal bids. This problem can be cast as a particular multiple-choice knapsack problem (MCKP) with a nonlinear objective function. 
Drawing inspiration from the operations research literature~\citep{kellerer2004multiple}, we designed a heuristic algorithm which can efficiently compute a near-optimal bidding strategy when provided with data about the whole advertisement campaign.
We then trained an auto-bidding agent to imitate the oracle's bids, without access to privileged information from the future (Fig.~\ref{fig:overview}). We term this learning framework \textbf{Oracle Imitation Learning (OIL)}. 
We provide experimental evidence that this method outperforms online and offline RL and LP methods in sample efficiency and final performance.

In summary, our contributions are threefold:

1. \textbf{MCKP formulation:}  We model optimal advertisement bidding as a stochastic MCKP with a nonlinear objective and derive a deterministic approximation.

2. \textbf{Oracle imitation learning (OIL):} We introduce an  greedy algorithm (oracle) to efficiently solve the nonlinear MCKP and we design an imitation learning framework where an auto-bidding agent learns to imitate the bids of the oracle.

3. \textbf{Experimental validation.}  Using a simulator based on the AuctionNet dataset~\citep{su2024auction}, we show that OIL-trained agents achieve near-oracle performance and surpass other approaches, including PPO, BC, IQL, and LP.

\section{Related work}

Bid optimization has been thoroughly studied both in industry and academia~\citep{feldman2007budget,hosanagar2008optimal,ghosh2009adaptive,perlich2012bid,zhang2014optimal}. Early research focused on sponsored search~\citep{mehta2007adwords,edelman2007internet} or on using contextual information to optimize advertisement bids. However, with the popularity of RTB~\citep{muthukrishnan2009ad} and the progressively wider availability of user information, more recently targeted advertisement has become the predominant framework in online advertising~\citep{wang2017display}.

\textbf{Optimal bidding in online advertising.} Different approaches have been experimented, with varying popularity over the years. \citeauthor{ou2023survey}~\citep{ou2023survey} show that rule-based methods~\citep{geyik2016joint,lin2016combining,zhang2016feedback} have been replaced over time by prediction methods coupled with optimization algorithms~\citep{bompaire2021causal,gligorijevic2020bid,aggarwal2019autobidding}, while reinforcement learning (RL) has recently become more common~\citep{wang2017ladder,zhao2018deep,lu2019reinforcement,he2021unified}. Different works focus on specific aspects of online advertising, such as single constraint~\citep{cai2017real,wu2018budget}, multiple constraints~\citep{tang2020optimized,he2021unified,wang2022roi} or multiple agents~\citep{jin2018real,wen2022cooperative}. 
In this work, we focus on the problem of a single agent bidding in second-price auctions, with a single optimization target and a budget constraint. The target incorporates the trade-off between acquisition volume (number of conversions) and cost-per-acquisition (CPA), as proposed by~\citeauthor{xu2024auto}~\citep{xu2024auto}.

\textbf{Auto-bidding in second price auctions.}
A core component of advertisement bidding is estimating the utility value $V = K \times \mu$ of each IO, where $K$ is the target CPA and $\mu$ is the acquisition rate~\citep{xu2016lift}. $K$ is a parameter selected by the advertiser, stating how much it is ready to pay for an acquisition, while $\mu$ can be a function of the advertiser, the potential customer and the context~\citep{yan2009much,wu2009probabilistic,yuan2013real,zhang2014optimal}. It can be either estimated by the advertisement platform or by the advertiser itself~\citep{diemert2017attribution}. $V$ defines the \emph{fair value} of an IO~\citep{xu2016lift}. Theoretical results prove that bidding equal to the fair value (truth-telling) is optimal for second-price auctions, under the assumption of infinite budget~\citep{krishna2009auction,roughgarden2010algorithmic}. However, when the budget is limited, it might be convenient to bid below the fair value, to increase the acquisition efficiency~\citep{gong2023mebs}, or even above the fair value, if the advertiser can accept exceeding the target CPA to increase the number of acquisitions. The problem is more complex in the presence of multiple advertisement slots for a single IO. In this case, the efficiency of these slots varies with the position, and techniques such as bid shading can be employed to maximize the number of acquisitions~\citep{zhang2021meow,gong2023mebs}. Previous work has focused on optimal advertisement bidding when the fair value of an IO is unknown~\citep{ren2017bidding,gligorijevic2020bid,bompaire2021causal}. In this work, instead, we consider the framework defined by \citeauthor{su2024auction}~\citep{su2024auction}, in which the advertisement platform provides an estimation of the acquisition probability, and the advertiser optimizes its bids accordingly, as a function of its budget, target CPA and its understanding of the advertisement traffic patterns. We chose this specific problem because it can be tackled using the recently released AuctionNet dataset, which, with its more than 500 million recorded bids~\citep{su2024auction}, is the largest open-source advertisement bidding dataset to this date.

\textbf{Multiple-choice knapsack problem}
The MCKP is a generalization of the classic knapsack problem (KP), widely studied in the operations research literature~\citep{kellerer2004multiple}. In a KP, items of different value and different volume can be selected until there is no more space inside the knapsack. The MCKP adds the complication of grouping the items in classes and constraining the choice to at most one item per class (some formulations prescribe exactly one item per class~\citep{kellerer2004multiple}). The problem belongs to the NP-hard complexity class. Several algorithms have been proposed, including dynamic programming, convex hull, Lagrangian relaxation, approximate and exact methods~\citep{zemel1980linear,kellerer2004multiple,bednarczuk2018multi,caserta2019robust}. Finding the optimal IOs to bid for in an advertisement campaign is an example of a MCKP, where the budget corresponds to the size of the knapsack and the conversion probability is the value of an IO. The multiple-choice component is due to the presence of more than one slot per IO, and each advertiser can acquire at most one of them.

\section{Dataset and simulator}
\begin{figure}
    \centering
    \includegraphics[width=\linewidth]{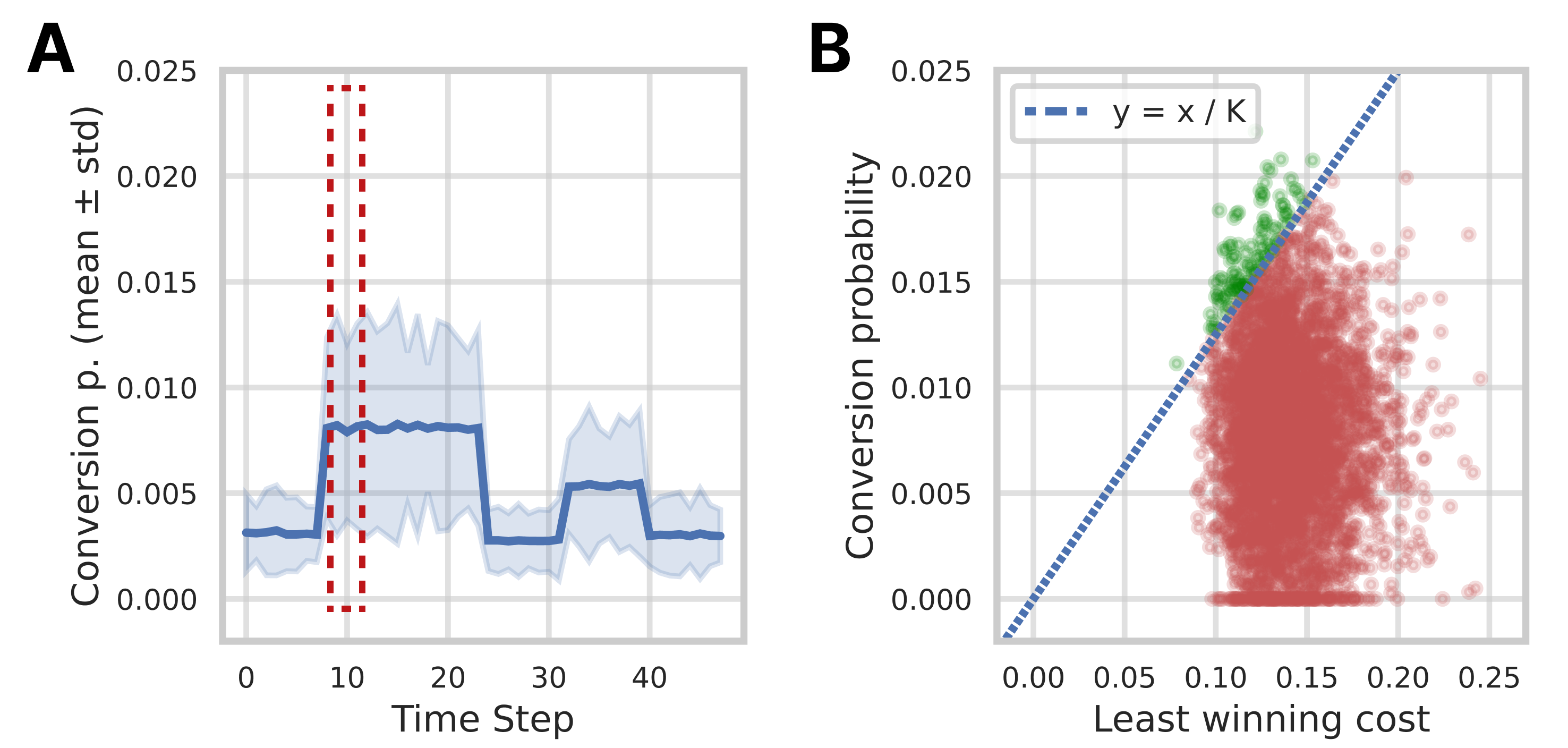}
    \caption{Example IOs from the AuctionNet-dense dataset. \textbf{A.} Conversion probabilities for a single advertiser and advertisement campaign. \textbf{B.} Conversion probability at one time step (red dotted line, A) as a function of their lowest price (3rd slot). IOs above the dotted line have an expected CPA lower than the target ($K=8$).}
    \label{fig:dataset}
\end{figure}
Open-source bidding datasets have historically been limited in size and feature diversity~\citep{ou2023survey}, and usually focus on predicting the click and conversion probability~\citep{diemert2017attribution}. Consequently, no common benchmark for auto-bidding agents exists, and recent works have relied on private datasets and closed-source algorithm implementations~\citep{wang2017ladder,zhang2021meow,zhou2021efficient,he2021unified,wang2022roi,wen2022cooperative}. This tendency complicates the comparison between methods~\citep{ou2023survey}. However, recently Alimama has released a large scale dataset with more than 500 million advertisement bids showcasing realistic advertisement traffic patterns~\citep{su2024auction}.

\textbf{Dataset.} The dataset comprises two datasets, \emph{AuctionNet-dense} and \emph{AuctionNet-sparse}, spanning 21 campaign periods of 48 time steps each. 
For each time step and IO, bids from 48 advertisers are recorded, competing for three advertisement slots assigned via second-price auction. This means that the advertisers presenting the three highest bids win the three slots, paying a price equal to the next highest bid (second to fourth). The main distinction between the two datasets is the average conversion probability, which is one order of magnitude lower in the sparse dataset, resulting in fewer conversions (hence the name "sparse").
Each advertiser is characterized by a fixed total campaign budget and a fixed target CPA, and bids according to its specific strategy. The dataset also provides the expected conversion probability for each IO, that was made available to the advertisers by the platform at bidding time. Conversion probabilities differ across advertisers for the same IO, reflecting varying product appeal to different potential customers. Advertisers face stochastic advertisement traffic (Fig.~\ref{fig:dataset} A).
To avoid both overbidding (and missing out on valuable IOs later due to lack of budget) and underbidding (and ending up with unspent budget that they had reserved for valuable IOs which never materialized), the advertisers need to adjust their strategy to the advertisement traffic they expect in the future. The conversion probabilities of the IOs at the same time step and for the same advertiser present a fairly spread distribution (Fig.~\ref{fig:dataset} B), with only a fraction being convenient for the advertiser (i.e., the bid necessary to win the IO leads to a lower CPA than the target).

\begin{figure}[t]
    \centering
    \includegraphics[width=\linewidth]{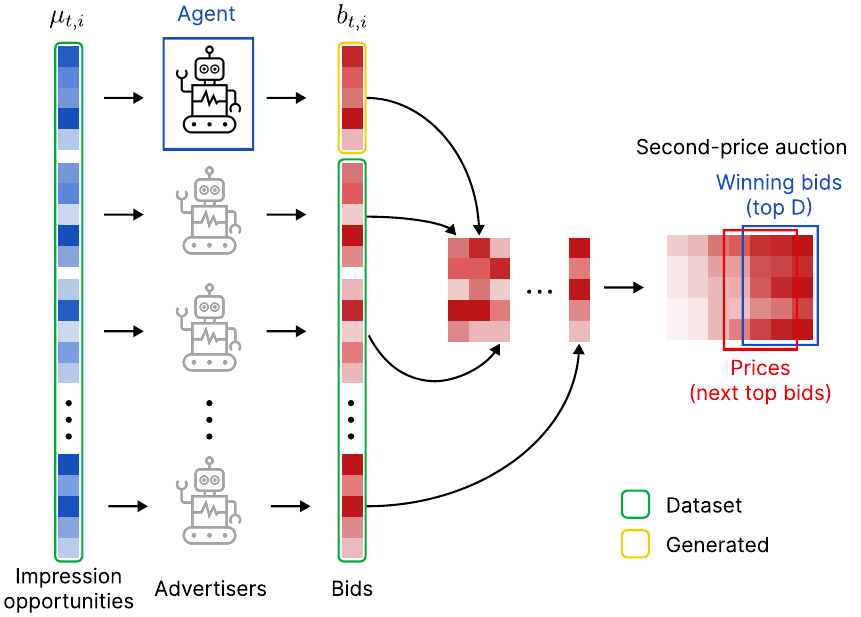}
    \caption{RTB simulator. At each step, the agent impersonates one of the advertisers and observes its IOs, replacing the bids of the impersonated advertiser with its own bids, while the bids of the other advertisers are kept unchanged. Slots and prices are decided according to a second-price auction.}
    \label{fig:spa}
\end{figure}

\textbf{Simulator.} To train auto-bidding agents, we developed an advertisement campaign simulator based on the AuctionNet dataset (Fig.~\ref{fig:spa}). The process is as follows: 1) the auto-bidding agent impersonates one of the 48 advertiser, sampled at random, and is assigned a budget and a target CPA. 2) At each time step, the simulator provides estimated conversion probabilities for available IOs (typically 1000 10,000 per time step). 3) The auto-bidding agent places bids without observing those of the competitors (closed-envelope auction). 4) Competitor bids are extracted from the dataset, assuming that they are not influenced by the auto-bidding agent's behavior. 5) Slot allocation and pricing are determined via second-price auction.

\section{Oracle policy}
The core of OIL is an oracle policy designed to identify the optimal set of slots to maximize acquisitions (e.g., clicks, conversions), given the entire information about the advertisement campaign. This problem can be modeled as a MCKP with nonlinear objective. By retrospectively analyzing campaign data (costs and conversion probabilities), the oracle determines how to improve the agent's actions. Unlike reward-based methods, which face challenges dealing with sparse conversions and noisy estimates, this approach offers robust supervision by leveraging perfect information in hindsight.

\subsection{MCKP formulation}
The problem of finding the optimal set of slots can be formulated as the following nonlinear stochastic integer optimization problem:
\begin{align}
\text{maximize} \quad & U =\,\, \mathbb{E} \left[ \min\left(1, \left( K \frac{\sum_{t,i,d} A_{t,i,d} x_{t,i,d}}{\sum_{t,i,d} C_{t,i,d} x_{t,i,d}} \right)^2 \right) \sum_{t,i,d} A_{t,i,d} x_{t,i,d} \right] \nonumber\\
\text{subject to} \quad & \sum_{t,i,d} C_{t,i,d} x_{t,i,d} \leq B, \nonumber \\
& \sum_d x_{t,i,d} \leq 1, \quad \forall t, i, \qquad x_{t,i,d} \in \{0, 1\},
\label{eq:stochastic}
\end{align}
where
\begin{itemize}
    \item $x_{t,i,d}$: A binary variable indicating the outcome of a bid. $x_{t,i,d} = 1$ if the agent wins the slot $d$ for impression $i$ at time step $t$, and $x_{t,i,d} = 0$ otherwise.
    
    \item $A_{t,i,d}$: A Bernoulli random variable representing an acquisition event, with mean $\beta_{t,i,d}$, which indicates the probability of acquiring a conversion upon winning the slot.
    
    \item $C_{t,i,d}$: The cost of slot $d$ for impression $(t, i)$. It is defined as $C_{t,i,d} = k_{t,i,d} \, H_{t,i,d}$,
    where $k_{t,i,d}$ is the cost of the slot (as determined by the second price auction), and $H_{t,i,d}$ is a Bernoulli random variable with parameter $h_d$ representing the probability of the ad being exposed (exposure probability), which only depends on the slot position $d$).
    
    \item $B$: The total budget constraint for the campaign.
    
    \item $K$: The target cost-per-acquisition (CPA) constraint.

    \item $\beta_{t,i,d}$: The mean acquisition probability, a Gaussian random variable $\mathcal{N}(H_{t,i,d}\mu_{t,i}, \sigma_{t,i}^2)$ , clipped between 0 and 1. 
    $\mu_{t,i}$ and $\sigma_{t,i}$ are the mean and standard deviation of the conversion probability for impression $(t, i)$.
\end{itemize}

The objective $U$, proposed by~\citeauthor{xu2024auto}\citep{xu2024auto} is the product of the total acquisitions $\sum_{t,i,d} A_{t,i,d} x_{t,i,d}$ and a penalty coefficient $\left(1, \left( K \frac{\sum_{t,i,d} A_{t,i,d} x_{t,i,d}}{\sum_{t,i,d} C_{t,i,d} x_{t,i,d}} \right)^2 \right)$, which is smaller than $1$ when $\frac{\sum_{t,i,d} C_{t,i,d} x_{t,i,d}}{\sum_{t,i,d} A_{t,i,d} x_{t,i,d}}$ (the CPA) exceeds $K$ (the target CPA).

Using a large number approximation for the objective and replacing the budget constraint by its expected value, we obtain the following:

\begin{align}
\text{maximize} \quad & U =\,\, \min\left(1, \left( K \frac{\sum_{t,i,d} h_d \mu_{t,i}  x_{t,i,d}}{\sum_{t,i,d} h_d k_{t,i,d}  x_{t,i,d}} \right)^2 \right) \left(\sum_{t,i,d} h_d \mu_{t,i}  x_{t,i,d} \right) \nonumber \\
\text{subject to} \quad & \sum_{t,i,d} h_d k_{t,i,d} x_{t,i,d} \leq B, \nonumber \\
& \sum_d x_{t,i,d} \leq 1, \quad \forall t, i, \qquad x_{t,i,d} \in \{0, 1\}
\label{eq:simplified}
\end{align}

(detailed derivation in the Appendix). We remark that, during the simulation, the budget constraint is enforced exactly, and the constraint in expectation is only used to find a solution to the problem.

\subsection{Heuristic algorithm}
Exact solutions to Problem~\ref{eq:simplified} are computationally infeasible for campaigns involving hundreds of thousands of IOs, such as the ones in the AuctionNet dataset.
Instead, we employ a greedy heuristic approximation. Inspired by knapsack problem solutions, the algorithm ranks slots by efficiency (e.g., conversion probability over cost, $\mu_{t,i}/k_{t,i,d}$) and selects slots until the budget is exhausted. Expected conversion probabilities and costs are adjusted for exposure probabilities since conversions and costs only occur when ads are displayed (effective conversion probability $\tilde{\mu}_{t,i,d} = \mu_{t,i} h_d$ and effective cost $\tilde{k}_{t,i,d} = k_{t,i,d} h_d$). This ranking approach leads to an efficient utilization of the budget and controls the CPA (which is, in fact, a measure of efficiency), but the multiple-choice nature of the problem adds complexity. Algorithm~\ref{algo:heruistic_oracle} maintains the structure of standard knapsack heuristics~\citep{dantzig2002linear,kellerer2004multiple}, with two main blocks: advertisement slot ranking by efficiency (lines 3-12) and accumulation of slots until the budget is over (lines 14-24). However, it presents three key modifications:

\textbf{1. Efficiency measure.} Slots are ranked by a generic efficiency function $f(k_{t,i,d}, \mu_{t,i}, \Delta c_{t,i,d}, \Delta m_{t,i,d})$, depending on the cost $k_{t,i,d}$, the conversion probability $\mu_{t,i}$, the slot upgrade cost $ \Delta c_{t,i,d} = \tilde{k}_{t,i,d} - \tilde{k}_{t,i,d+1}$ and slot upgrade conversion probability $ \Delta m_{t,i,d} = \tilde{\mu}_{t,i,d} - \tilde{\mu}_{t,i,d+1}$.
Importantly, the ranking function must give higher score to slots with higher index within the same IO, as this property allows to deal with multiple slots.

\textbf{2. Handling multiple slots.} The constraint $\sum_d x_{t,i,d} \leq 1$ imposes that an advertiser can win at most one slot per IO. This means that, if slot $d$ of IO $(t,i)$ is ranked before slot $d'$ of the same IO, slot $d$ needs to be removed from the solution set before acquiring slot $d'$. This action can be modeled by subtracting the cost and expected conversions of slot $d$ and adding those of slot $d'$. If the ranking function guarantees that the slots are sorted in descending order (i.e., slot $d+1$ is always more efficient than slot $d$), then the algorithm can pre-compute the cost and conversion upgrades (lines 5 and 6) and use them to efficiently derive the cumulative conversions and costs (lines 18 and 19). 

\textbf{3. Nonlinear objective function.} Unlike linear knapsack problems, acquiring slots whose efficiency is lower than $1/K$ can lower the overall score, because of a decrease in the coefficient $K/CPA$. Therefore, Algorithm~\ref{algo:heruistic_oracle} computes the score every time a new slot is acquired (line 20) and chooses $R^*$ so that the score is maximized.

The complexity of Algorithm~\ref{algo:heruistic_oracle} is $\mathcal{O}(DN\log(DN))$, where $N$ is the number of IOs, with $D$ slots each, due to the sorting of all the slots of the campaign (line 10).


\begin{algorithm}[t]
\caption{Oracle Algorithm}
\begin{algorithmic}[1]
\item[]
\begin{minipage}{\linewidth}
    \centering
    \includegraphics[width=0.8\linewidth]{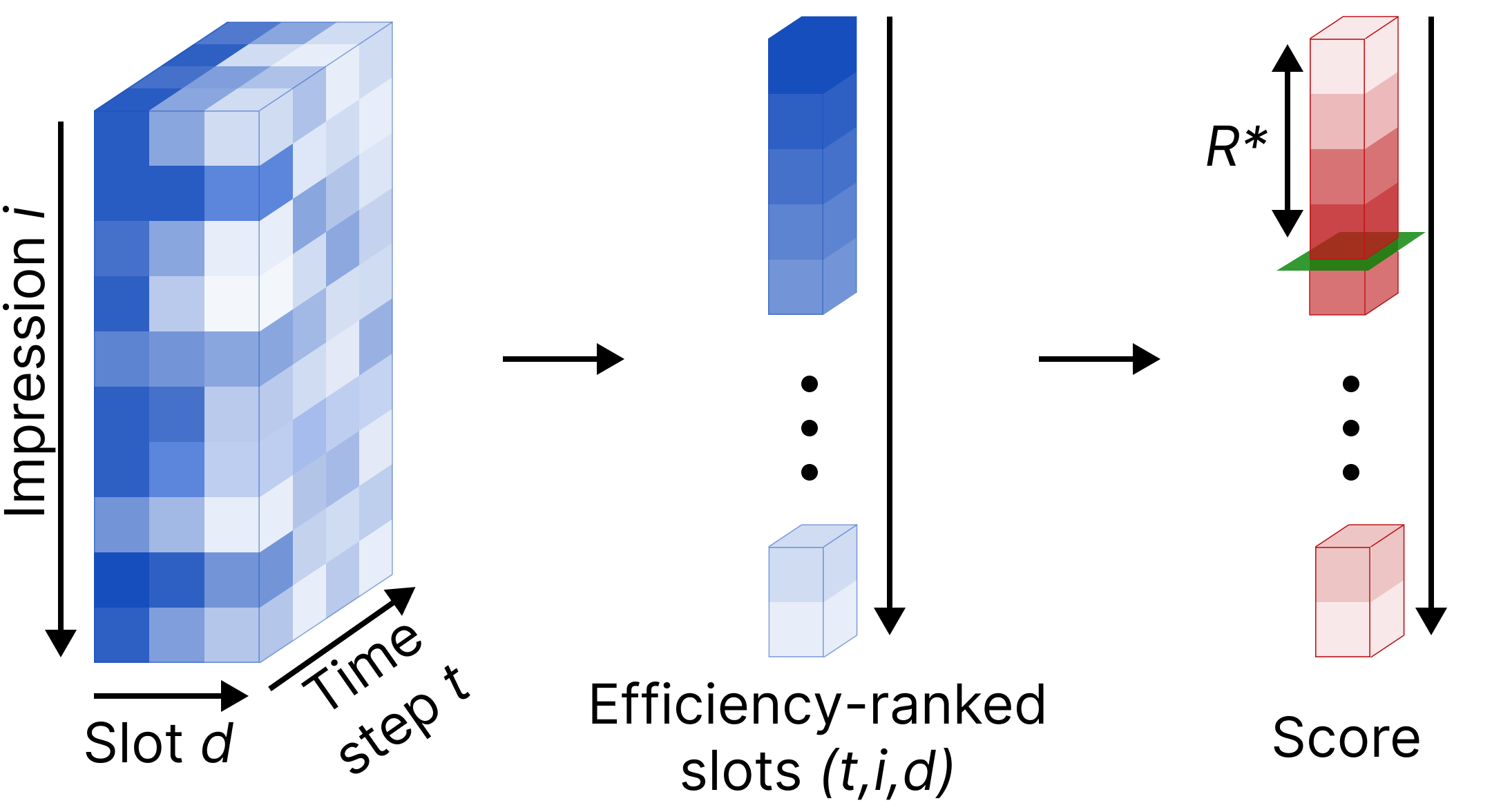}
\end{minipage}
\STATE Provided: expected conversion probabilities $\mu_{t,i}$, expected exposure probability $h_d$ and expected costs $c_{t,i,d}$ for ad impression $i$, slot $d$ and time step $t$ current total cost $tot\_cost_s$, current total conversions $tot\_conv_s$, target CPA $K$, budget $B$, $T$ campaign duration, $M(t)$ number of IOs at time $t$, $D$ slots.
\STATE \textbf{Compute ranked impression slots}
\IF{$s = 1$}
    \FOR{$t \in \{1, ..., T\}$, $i \in \{1, ..., M(t)\}$, $d \in \{1, ..., D\}$}
        \STATE $\Delta c_{t,i,d} = k_{t,i,d} h_d - k_{t,i,d+1}h_{d+1}$ (slot cost upgrade)
        \STATE $\Delta m_{t,i,d} = \mu_{t,i} (h_d - h_{d+1})$ (slot conversion upgrade)
        \STATE [NOTE: we consider $h_{D+1} = k_{t,i,D+1} = 0$]
        \STATE $\eta_{t,i,d} = f(k_{t,i,d}, \mu_{ti}, \Delta c_{t,i,d}, \Delta m_{t,i,d})$ (efficiency measure)
    \ENDFOR
    \STATE Compute $R=argsort(\eta_{t,i,d})$ by decreasing efficiency
    \STATE Store $R$ for future use
\ENDIF

\STATE \textbf{Compute expected cumulative conversions, cost, CPA and score acquiring the top-$j$ ranked slots}
\STATE Update $R = \{(t,i,d): (t,i,d) \in R \land t \geq s\}$ preserving order
\STATE $j = 1$, $cum\_conv_{-1} = tot\_conv_s$, $cum\_cost_{-1} = tot\_cost_s$
\FOR{$(t,i,d) \in R$}
    \STATE $cum\_conv_j = cum\_conv_{j-1} + \Delta m_{t,i,d}$
    \STATE $cum\_cost_j = cum\_cost_{j-1} + \Delta c_{t,i,d}$
    \STATE $CPA_j = cum\_cost_j / cum\_conv_j$
    \STATE $score_j = \min\left(1, \left(\frac{K}{CPA_j}\right)^2\right) cum\_conv_j$
    \IF{$cum\_cost_j > B$}
        \STATE break (no need to compute past the budget)
    \ENDIF
    \STATE $ j = j + 1$
\ENDFOR
\STATE \textbf{Determine the list of slots to acquire}
\STATE $j^* = \argmax_j score_j$ (least efficient slot to win)
\RETURN $R^* = \{(t,i,d)_j \in R, j<j*\}$

\end{algorithmic}
\label{algo:heruistic_oracle}
\end{algorithm}

\subsection{Oracle-slot algorithm}
\label{sec:slot}
Here we describe the first version of Algorithm~\ref{algo:heruistic_oracle}, where slots are ranked by their efficiency.

\begin{definition}
    \textbf{D-slot second-price IO (D-SPIO).} A D-SPIO $\mathcal{I} = \{c_1, ..., c_D, p_1, ..., p_D\}$ is a collection of $D\in \mathbb{N}$ expected costs $c_d \in \mathbb{R}^+$ and $D$ expected conversion probabilities $p_d \in [0, 1]$ which admits $\delta_d \in (0,1]$ cost discount factors and $h_d \in (0,1]$ exposure discount factors ($d \in \{1, ..., D\}$) such that:
    \begin{enumerate}
        \item $c_d = k_1 \delta_d h_d \qquad \forall d \in \{1, ..., D\}$ 
        \item $p_d = \mu h_d \qquad \forall d \in \{1, ..., D\}$ 
        \item $ 0 < h_i \leq h_j \leq 1  \qquad \forall i, j \in \{1, ..., D\}, i>j$ 
        \item $ 0 < \delta_i < \delta_j \leq 1  \qquad \forall i, j \in \{1, ..., D\}, i>j$ 
    \end{enumerate}
    with $k_1\in \mathbb{R}$ being the cost of the first slot (before exposure discount) and $\mu \in \mathbb{R}$ the conversion probability (before exposure discount).
\end{definition}

\begin{remark}
    The definition reflects a multi-slot IO, where slots are assigned according to an auction, under the assumption that slots with a higher index have a smaller probability of being displayed to a customer, and that the slots are assigned from the first to the last to advertisers ranked from the highest bid to the D-th top bid. In this case, the expected cost of a slot is given by the cost $k_1$ of the first slot, discounted by the exposure probability (if an advertisement is not exposed, no price is paid) and by the fact that later slots have a lower base price. Also the conversion probability $\mu$ is discounted by the exposure probability, because an ad that is not exposed cannot lead to a conversion. 
\end{remark}

\begin{definition}
    \textbf{Slot efficiency.} The efficiency of slot $d$ of an D-SPIO is $\eta_n = \frac{p_d}{c_d}$, i.e., the ratio between the conversion probability and the cost of that slot.
\end{definition}
\begin{remark}
    The slot efficiency of slot $d$ is also equal to $\frac{\mu}{\delta_d k_1}$, which means that it is independent of the exposure probability of the slot.
\end{remark}

The first version of Algorithm~\ref{algo:heruistic_oracle}, which we call \emph{oracle-slot} uses the efficiency measure
\begin{equation}
    f(\mu, k_{d}) = \frac{\mu}{k_{d}}.
\end{equation}

This efficiency measure is a natural choice when the objective is to maximize the number of conversions. It bears two important properties: (1) it decreases with the slot index, so it can be seamlessly implemented in Algorithm~\ref{algo:heruistic_oracle} (Lemma~\ref{lemma:sorted_efficiency}), and (2) the solution can be implemented by a bidding agent outputting bids proportional to the expected conversion probability $\mu_{t,i,d}$ of each impression (Lemma~\ref{lemma:const_bid_coef}). However, its solutions might be suboptimal in certain situations (Section \ref{sec:upgrade}). We now formalize these properties.

\begin{restatable}{lemma}{sortedEfficiency}
    \label{lemma:sorted_efficiency}
    \textbf{Slot efficiencies are sorted.} Given a D-SPIO, the efficiency of a slot monotonically increases with its position, i.e., $\eta_i > \eta_j$ if $i > j$.
\end{restatable}

\begin{proof}
    See appendix.
\end{proof}

\begin{restatable}{lemma}{constBidCoef}
    \label{lemma:const_bid_coef}
    \textbf{A constant bidding coefficient identifies the oracle-slot's solution.} Consider the set $R^*$ of IOs output by Algorithm~\ref{algo:heruistic_oracle} and assume that no two advertisement slots have exactly the same efficiency $\eta_{t,i,d}$. Then a bidding agent whose bid for the IO $i$ at time $t$ is $b_{t,i} = \alpha \mu_{t,i}$, with $\alpha = \max{\{1/\eta_{t,i,d} \,\,s.t.\,\, (t, i, d) \in R^*\}}$ wins all and only the slots in $R^*$. If multiple slots of the same IO are part of $R^*$, the bidding agents wins the one with the smallest $d$.
\end{restatable}
\begin{proof}
    See Appendix.
\end{proof}

\begin{table}[h!]
    \centering
    \caption{Two examples of two impression opportunities (IOs) with two slots each. The only difference between the two examples is the conversion probability $\mu$ for the first IO, which doubles. It leads to different derived quantities, highlighted in red. In this example with only two slots, upgrading from slot 3 to slot 2 is equivalent to acquiring slot 2.}
    \resizebox{\linewidth}{!}{
        \begin{tabular}{llcccc}
            \hline
            \multirow{2}{*}{\textbf{Parameter}} & \multirow{2}{*}{\textbf{Symbol}} & \multicolumn{2}{c}{\textbf{Example 1}} & \multicolumn{2}{c}{\textbf{Example 2}} \\
            \cmidrule(lr){3-4} \cmidrule(lr){5-6}
            & & \textbf{IO 1} & \textbf{IO 2} & \textbf{IO 1} & \textbf{IO 2} \\
            \hline
            Exposure p. slot 1 & $\bm{h_1}$ & 1.000 & 1.000 & 1.000 & 1.000 \\
            Exposure p. slot 2 & $\bm{h_2}$ & 0.800 & 0.800 & 0.800 & 0.800 \\
            Cost slot 1 & $\bm{k_1}$ & 1.000 & 1.000 & 1.000 & 1.000 \\
            Cost slot 2 & $\bm{k_2}$ & 0.375 & 0.875 & 0.375 & 0.875 \\
            Conversion p. & $\bm{\mu}$ & 0.100 & 0.040 & \textcolor{red}{0.200} & 0.040 \\
            \midrule
            Mean cost slot 1 & $\bm{\tilde{k}_1 = k_1 h_1}$ & 1.000 & 1.000 & 1.000 & 1.000 \\
            Mean cost slot 2 & $\bm{\tilde{k}_2 = k_2 h_2}$ & 0.300 & 0.700 & 0.300 & 0.700 \\
            Mean conv. slot 1 & $\bm{\tilde{\mu}_1 = \mu h_1}$ & 0.100 & 0.040 & \textcolor{red}{0.200} & 0.040 \\
            Mean conv. slot 2 & $\bm{\tilde{\mu}_2 = \mu h_2}$ & 0.080 & 0.032 & \textcolor{red}{0.160} & 0.032 \\
            Efficiency slot 1 & $\bm{\eta_1 = \mu / k_1}$ & 0.100 & 0.040 & \textcolor{red}{0.20} & 0.040 \\
            Efficiency slot 2 & $\bm{\eta_2 = \mu / k_2}$ & 0.267 & 0.046 & \textcolor{red}{0.533} & 0.046 \\
            Eff. upgrade 2 $\rightarrow$ 1 & $\bm{\tilde{\eta}_{1} = \frac{\tilde{\mu}_1 - \tilde{\mu}_2}{\tilde{k}_1 -\tilde{k}_2}}$ & 0.029 & 0.027 & \textcolor{red}{0.057} & 0.27\\
            Eff. upgrade 3 $\rightarrow$ 2 & $\bm{\tilde{\eta}_{2} = \eta_2}$ & 0.267 & 0.046 & \textcolor{red}{0.533} & 0.046 \\
            \hline
        \end{tabular}
    }
    \label{tab:example}
\end{table}

\subsection{Oracle-upgrade algorithm}
\label{sec:upgrade}
The oracle-slot algorithm suffers from one limitation: maybe counter-intuitively, adding the most efficient slot currently not in the solution set $R^*$ is not always the most efficient way to extend $R^*$. We clarify this statement with an example. For simplicity, let us consider two IOs with two advertisement slots each (D=2). We call $S_{i,j}$ the slot $j$ of IO $i$. The specifications of the two IOs are detailed in Table~\ref{tab:example}. According to the efficiency values of Example 1, we have that $S_{1,2}$ is the most efficient, with $\eta_{1,2} = 0.278$, followed by $S_{1,1}$ ($\eta_{1,1} = 0.100$), $S_{2,2}$ ($\eta_{2,2} = 0.048$) and $S_{2,1}$ ($\eta_{2,1} = 0.040$). And in fact, when bidding proportionally to the conversion probability, we would win slot $S_{1,2}$ with a bid coefficient $\alpha$ such that  $1/\eta_{1,2} \leq \alpha < 1/\eta_{1,1}$, $S_{1,1}$ with $1/\eta_{1,1} \leq \alpha < 1/\eta_{2,2}$, $S_{1,1}$ and $S_{2,2}$ with $1/\eta_{2,2} \leq \alpha < 1/\eta_{2,1}$ and finally $S_{1,1}$ and $S_{2,1}$ with $\alpha \geq 1/\eta_{2,1}$. However, we can easily verify that this strategy does not always achieve the maximum expected conversions for a given budget. In fact, winning $S_{1,1}$ leads to $0.1$ expected conversions for an expected cost of $1$, while winning $S_{1,2}$ and $S_{2,2}$ leads to $0.112$ expected conversions, for the same expected cost. This inefficiency is caused by $S_{1,2}$ being much more efficient than $S_{1,1}$. Therefore, although $S_{1,1}$ is more efficient than $S_{2,2}$, it is still better to win $S_{2,2}$ than $S_{1,1}$, in order not to lose the very efficient $S_{1,2}$ (reminder: we cannot win both $S_{1,1}$ and  $S_{1,2}$, because they are slots of the same IO).

From the previous example, one might conclude the best strategy is to always bid for the most efficient slot of each IO. However, this intuition is incorrect. Let us consider another two IOs (Table~\ref{tab:example}, Example 2). The features are identical to the previous example, except for the conversion probability of impression 1, which are doubled. In this case, we can verify that winning $S_{1,1}$ leads to $0.2$ expected conversions for an expected cost of $1$, while winning $S_{1,2}$ and $S_{2,2}$ leads to $0.192$ expected conversions, for the same expected cost. In this case, \emph{upgrading} from slot 2 to slot 1 in IO 1 is better than acquiring slot 2 of IO 2. But how can we frame the problem so that we can rank "actions" in order of efficiency and perform one action after the other, obtaining the most efficient combination of bids? We answer this question next.
 
We interpret the problem of finding the optimal set of bids as sequential decision making. The first decision is simple: if we have enough budget, the most efficient slot is $S_{1,2}$. At this point, we face the following dilemma: is it better to upgrade to slot $S_{1,1}$, giving up slot $S_{1,2}$, or to keep slot $S_{1,2}$ and buy slot $S_{2,2}$? In order to take the best decision, we should not compare the efficiency of $S_{2,2}$ and $S_{1,1}$ (we have seen that it can lead to suboptimal solutions), but instead we should compare the efficiency of $S_{2,2}$ and the efficiency of the upgrade from $S_{1,2}$ to $S_{1,1}$.

\begin{definition}
    \textbf{Slot upgrade efficiency.} The efficiency of upgrading from slot $j$ to slot $i$ of the same D-SPIO (i < j) is 
    \begin{equation}
        \tilde{\eta}_{i,j} = \frac{p_i - p_j}{c_i - c_j} = \frac{\mu(h_{i} - h_{j})}{k_1( \delta_i h_{i} - \delta_{j} h_{j})},
    \end{equation}
    i.e., the ratio between the expected difference in conversion probability and the expected difference in cost when winning slot $i$ instead of slot $j$.
\end{definition}
\begin{remark}
    The exposure probability is essential in defining a meaningful slot upgrade efficiency. Without factoring in the exposure probability, the efficiency of an upgrade would be $0$, because it would lead to an increased cost for the same conversions.
\end{remark}
\begin{remark}
    The ordering valid for slot efficiency does \textbf{not} hold for the slot upgrade efficiency, i.e., there can be cases in which $\tilde{\eta}_{i, k} > \tilde{\eta}_{k,j}$ for $i < k < j$.
\end{remark}

In the examples of Table~\ref{tab:example}, comparing the efficiency of the upgrade to the efficiency of acquiring a new slot leads us to the correct decision: in the Example 1, we have $\tilde{\eta}_{1,1} = 0.029$, which is less than $\eta_{2,2} = 0.04$. In Example 2, instead, $\tilde{\eta}_{1,1} = 0.057$, while $\eta_{2,2} = 0.04$, meaning that upgrading to slot 1 is more efficient in this case. Following these considerations, we define the efficiency function of the oracle-upgrade as:
\begin{equation}
    f(\Delta c_{t,i,d}, \Delta m_{t,i,d}) = \frac{\Delta m_{t,i,d}}{\Delta c_{t,i,,d}},
\end{equation}
where $\Delta m_{t,i,d} = \mu_{t,i}(h_{d} - h_{d'})$ and $\Delta c_{t,i,,d} = k_{t,i,d} h_d - k_{t,i,d'} h_{d'}$.
Here $d'>d$ indicates the slot that is being upgraded to $d$, and depends on the slot $(t,i,d)$. In fact, while consecutive upgrades are not necessarily sorted by efficiency, skipping specific slots when upgrading can restore the order. Lemmas~\ref{lemma:d_best} and \ref{lemma:harmonic} clarify how we can construct a sequence of upgrades from slot $D$ to slot $1$ of the same IO with monotonically decreasing efficiency, as required by the oracle algorithm.

\begin{restatable}{lemma}{dBest}
    \label{lemma:d_best}
    \textbf{The last slot $D$ is more efficient than any upgrade.} Given a D-SPIO $\mathcal{I} = \{c_d, p_d, \mu, k_1, \delta_d\}$, we have that $\eta_D > \tilde{\eta}_{i,j}$, for any $i, j \in \{1, ..., D\}$ s.t. $i < j$.
\end{restatable}

\begin{proof}
    See Appendix.
\end{proof}

\begin{restatable}{lemma}{harmonic}
    \label{lemma:harmonic}
    \textbf{The efficiency of an upgrade is the weighted harmonic mean of the efficiencies of intermediate upgrades.} Given a D-SPIO $\mathcal{I} = \{c_d, p_d, \mu, k_1, \delta_d\}$, we consider three upgrades $\tilde{\eta}_{i, k}$, $\tilde{\eta}_{k,j}$ and $\tilde{\eta}_{i,j}$, with $i \leq k \leq j$. Then
    \begin{equation}
        \frac{1}{\tilde{\eta}_{i,j}} = \beta \frac{1}{\tilde{\eta}_{i,k}} + (1-\beta) \frac{1}{\tilde{\eta}_{k,j}},  
    \end{equation}
    with $\beta = \frac{h_i - h_k}{h_i - h_j}$, which implies that $\tilde{\eta}_{i,j} \leq \min ( \tilde{\eta}_{i,k}, \tilde{\eta}_{k,j})$ .

\end{restatable}

\begin{proof}
    See Appendix.
\end{proof}
\begin{remark}
    Lemmas~\ref{lemma:d_best} and~\ref{lemma:harmonic} provide a way to define upgrades with decreasing efficiency. Slot $D$ can always be included in the ranking, as it is more efficient than any upgrade (Lemma~\ref{lemma:d_best}). Subsequent upgrades are added in order. If upgrade $i$ is more efficient than upgrade $i-1$, they are merged into a new upgrade (whose efficiency is computed according to Lemma~\ref{lemma:harmonic}). The new upgrade might also be more efficient than the previous one, case in which another merge is performed, until we get monotonically decreasing sequence of efficiencies. The worst-case complexity of this algorithm is $D^2$, which does not impact the overall complexity of the oracle if $D$ is much smaller than the number of IOs.
\end{remark}

\begin{remark}
     Differently from the case of the oracle-slot, in general there is no constant bidding coefficient $\alpha$ to win all and only the slots selected by the oracle-upgrade. In Example 1 of Table~\ref{tab:example}, one cannot win only $S_{1,2}$ and $S_{2,2}$ with a constant coefficient $\alpha$, because $S_{2,2}$ is less efficient than $S_{1,1}$. In fact, to improve over the oracle-slot, the oracle-upgrade sacrifices placing bids that are linearly proportional to the conversion probability. As we will see in Section~\ref{sec:experiments}, imitating an agent which outputs a specific bid for each IO is more difficult and might not be the best solution.
\end{remark}

\begin{restatable}{theorem}{optimalityGap}
    \label{th:gap}
    \textbf{Optimality gap of the oracle-upgrade solution.} Let $x_{t,i,d} = \mathbbm{1}_{R^*}(t,i,d)$ be the solution of Problem~\ref{eq:simplified}, with $R^*$ computed according to the oracle-upgrade algorithm ($\mathbbm{1}_{R^*}(t,i,d)$ is the indicator function of the set $R^*$). If $\frac{\sum_{t,i,d}h_d k_{t,i}x_{t,i,d}}{\sum_{t,i,d}h_d \mu_{t,i}x_{t,i,d}} < K$, then the following bound for the optimality gap holds:
    \begin{equation}
        U_{\max} - U \leq \frac{\sum_{t,i,d}h_d \mu_{t,i}x_{t,i,d}}{\sum_{t,i,d}h_d k_{t,i}x_{t,i,d}} \Delta B,
    \end{equation}
    where $U_{\max}$ is the optimal score for Problem~\ref{eq:simplified} with budget $B$ and target CPA $K$ and $\Delta B$ is the remaining budget at the end of the campaign.
\end{restatable}

\begin{proof}
    See appendix.
\end{proof}
\begin{remark}
    The condition $\frac{\sum_{t,i,d}h_d k_{t,i}x_{t,i,d}}{\sum_{t,i,d}h_d \mu_{t,i}x_{t,i,d}} < K$ means that the CPA constraint must be satisfied for the inequality to hold. Experimental evidence shows that this is often the case (Section~\ref{sec:experiments}). If the CPA exceeds $K$, the objective becomes nonlinear, complicating the estimation of an optimality gap. Furthermore, our experiments also show that the oracle consumes 99\% of the budget on average, making the gap small in practice.
\end{remark}
\begin{table*}[h]
    \centering
    \caption{Performance of OIL and different baselines. First block: baselines using the offline dataset. Second block: baselines using the bidding simulator. Third block: OIL-slot. Fourth block: different oracles.
    The performance is reported as mean $\pm$ standard error, across 3 different training experiments with different random seed. The oracle policies do not require any training and only the average scores are reported.}
    \resizebox{\linewidth}{!}{%
    \begin{tabular}{lcccccccc}
        \toprule
        \multirow{2}{*}{\textbf{Algorithm}} & 
        \multicolumn{2}{c}{\textbf{Cost / Budget}} & 
        \multicolumn{2}{c}{\textbf{Target CPA / CPA}} & 
        \multicolumn{2}{c}{\textbf{Conversions}} & 
        \multicolumn{2}{c}{\textbf{Score}~($\uparrow$)} \\
        \cmidrule(lr){2-3}\cmidrule(lr){4-5} \cmidrule(lr){6-7} \cmidrule(lr){8-9}
         & Dense & Sparse & Dense & Sparse & Dense & Sparse & Dense & Sparse \\
        \midrule
        Online LP & $0.55 \pm 0.0001$ & $0.71 \pm 0.0002$ & $1.10 \pm 0.0002$ & $1.07 \pm 0.0001$ & $337.62 \pm 0.07$ & $25.94 \pm 0.005$ & $331.66 \pm 0.11$ & $24.07 \pm 0.02$ \\
        BC & $0.53 \pm 0.21$ & $0.58 \pm 0.06$ & $1.03 \pm 0.20$ & $1.14 \pm 0.06$ & $263.15 \pm 75.14$ & $22.19 \pm 0.71$ & $176.19 \pm 35.57$ & $19.13 \pm 0.29$ \\
        IQL-score & $0.80 \pm 0.04$ & $0.43 \pm 0.09$ & $0.98 \pm 0.04$ & $1.51 \pm 0.13$ & $473.69 \pm 17.82$ & $18.72 \pm 2.98$ & $383.35 \pm 5.92$ & $16.88 \pm 2.41$ \\
        IQL-shaped & $0.60 \pm 0.06$ & $0.41 \pm 0.01$ & $1.14 \pm 0.05$ & $1.40 \pm 0.02$ & $390.95 \pm 30.66$ & $18.05 \pm 0.44$ & $339.30 \pm 24.41$ & $15.95 \pm 0.67$ \\
        \midrule
        PPO-score & $0.82 \pm 0.02$ & $0.75 \pm 0.01$ & $0.91 \pm 0.02$ & $0.91 \pm 0.03$ & $493.63 \pm 3.12$ & $26.26 \pm 0.94$ & $381.97 \pm 10.49$ & $21.89 \pm 1.17$ \\
        PPO-shaped & $0.87 \pm 0.04$ & $0.89 \pm 0.02$ & $0.91 \pm 0.02$ & $1.03 \pm 0.02$ & $518.13 \pm 11.90$ & $33.28 \pm 0.16$ & $384.71 \pm 13.12$ & $28.33 \pm 0.30$ \\
        PPO-shaped-2s & $0.78 \pm 0.02$ & $0.88 \pm 0.01$ & $0.81 \pm 0.02$ & $1.07 \pm 0.01$ & $413.40 \pm 18.90$ & $34.43 \pm 0.41$ & $291.58 \pm 22.41$ & $29.95 \pm 0.36$ \\
        \midrule
        \textbf{OIL-slot (ours)} & ${0.99 \pm 0.003}$ & ${0.98 \pm 0.001}$ & ${0.98 \pm 0.001}$ & ${1.13 \pm 0.001}$ & $611.29 \pm 1.14$ & $39.93 \pm 0.04$ & $\bm{466.36 \pm 0.43}$ & $\bm{35.29 \pm 0.05}$ \\
        \midrule
        Oracle-slot & 0.97 & 1.00 & 1.03 & 1.17 & 626.53 & 41.42 & 503.54 & 37.04 \\
        Oracle-upgrade & 0.98 & 0.99 & 1.06 & 1.19 & 650.10 & 42.23 & 532.60 & 38.08 \\
        Oracle-upgrade-2s & 0.97 & 0.99 & 1.04 & 1.17 & 631.11 & 41.61 & 510.85 & 37.49 \\
        \bottomrule
    \end{tabular}
    }
    \label{tab:results}
\end{table*}

\begin{algorithm}[b]
\caption{Oracle Imitation Learning (OIL)}
\begin{algorithmic}[1]
\STATE Provided: bidding agent $\pi_\theta$, oracle $\Omega$, bidding environment $\mathcal{E}$, number of episodes $N$, episode length $T$.
\FOR{$i \in \{1, ..., N\}$}
    \STATE Sample budget $B$, target CPA $K$, advertiser $A$
    \FOR{$t \in \{1, ..., T\}$}
        \STATE Get campaign history $h_t$, future $f_t$ and  IOs $o_t$ from $\mathcal{E}$
        \STATE $\bm{b}_t = \pi_{\theta_{t}} (h_t, o_t)$ \quad (agent's bids)
        \STATE $\bm{\tilde{b}}_t = \Omega(h_t, o_t, f_t)$ \quad (oracle's bids)
        \STATE $\theta_{t+1} = \theta_{t} - \nabla || \bm{b}_t -\bm{\tilde{b}}_t||^2$
        \STATE Advance $\mathcal{E}$ with bids  $\bm{b}_t$
    \ENDFOR
\ENDFOR
\end{algorithmic}
\label{algo:oil}
\end{algorithm}

\section{Oracle Imitation Learning (OIL)}
OIL is implemented as a behavior cloning (BC) algorithm where a student policy (the auto-bidding agent) learns to replicate the actions of an expert policy (the oracle). Unlike standard BC setups, OIL features an asymmetry of information between the student and the expert: the oracle has perfect knowledge about the advertisement traffic for the entire campaign, while the student does not. This encourages the auto-bidding agent to extract patterns from the past observations and predict the future advertisement traffic to compensate for its lack of information.

OIL employs an \emph{online} trining strategy, i.e., the agent generates new experience by interacting with the environment rather than using a pre-collected dataset (Algorithm~\ref{algo:oil}). 

This approach mitigates the issue highlighted by~\citeauthor{ross2011reduction}~\citep{ross2011reduction}, which occurs when the agent deviates from the policies in the dataset, encountering states outside of the dataset's distribution.
By collecting transitions online, the training state distribution is more likely to match that encountered during deployment, enhancing robustness and generalization.
An important detail of OIL is the dynamic update of the oracle bids at every time step, rather than relying on the optimal set $R^*$ computed at the start of the campaign (Algorithm~\ref{algo:heruistic_oracle}). While this introduces computational overhead, it mitigates the effects of the environment's stochasticity and differences between the agent's and the oracle's bids. The set $R^*$ is near-optimal under the assumptions that the agent follows the oracle policy and that actual costs and conversions align with the expected values. Updating the oracle bids with real-time data ensures near-optimality throughout the campaign.

\section{Experiments and results}
\label{sec:experiments}
We trained auto-bidding agents using OIL in the advertisement campaign simulator built on the advertisement traffic from the AuctionNet dense and sparse datasets. Budget and target CPA values for each simulated campaign were extracted from uniform distributions encompassing the dataset ranges. To capture campaign dynamics and temporal patterns from past auctions, we engineered 60 features (see Appendix) and fed them into a three-layer fully connected network with 256 units per layer to generate bids. Each OIL training experiment involved 10 million environment interactions, which were sufficient for convergence (Fig.~\ref{fig:lc_oil_ppo}). Detailed hyperparameters are listed in the Appendix. We evaluated three versions of OIL:

\begin{table}[b]
    \centering
    \caption{Score of OIL with different oracles.}
    \begin{tabular}{lcccc}
        \toprule
        \textbf{Method} & 
        \multicolumn{2}{c}{\textbf{Dense}} & 
        \multicolumn{2}{c}{\textbf{Sparse}} \\
        \cmidrule(lr){2-3} \cmidrule(lr){4-5}
         & OIL & Oracle & OIL & Oracle \\
        \midrule
        Slot & $\bm{466.36 \pm 0.43}$ & 503.54 & $\bm{35.29 \pm 0.05}$ & 37.04 \\
        Upgrade 2s & $459.21 \pm 9.43$ & 510.85 & $34.91 \pm 0.06$ & 37.49 \\
        Upgrade & $434.87 \pm 6.50$ & \textbf{532.60} & $29.42 \pm 0.12$ & \textbf{38.08} \\

        \bottomrule
    \end{tabular}
    \label{tab:oil_vs_oil}
\end{table}

\textbf{1. OIL-slot.} The oracle algorithms is the oracle-slot (Section~\ref{sec:slot}). By Lemma~\ref{lemma:const_bid_coef}, bids in the form $\bm{b}_t = \alpha \bm{\mu}_t$ (proportional to the expected conversion probabilities) can win the advertisement slots in $R^*$ as output by the oracle algorithm (Algorithm~\ref{algo:heruistic_oracle}), when the slots are ranked by slot efficiency. Therefore, the auto-bidding agent outputs a scalar bidding coefficient, aiming to mimic the oracle's $\alpha$.

\textbf{2. OIL-upgrade.} The oracle algorithm is the oracle-upgrade (Section~\ref{sec:upgrade}). In this case, a constant bidding coefficient cannot secure the advertisement slots in $R^*$ (Section~\ref{sec:upgrade}). Thus, to enable the agent to output bids tailored for each IO, the observations were augmented with the conversion probability statistics ($\mu$ and $\sigma$), enabling the agent to regress the oracle's bids for individual IOs.

\textbf{3. OIL-upgrade-2s.} 
The oracle-upgrade is approximated using three coefficients, defining the slopes of two lines and their crossing point (two-slopes approximation). This approximation captures the shape of the optimal bid distribution, which tends to underbid for IOs with high expected conversion rate (see Appendix). This pattern is similar to bid shading, which is a common strategy in first-price auctions, but which - to the best of our knowledge - has not been observed in second-price auctions. The reason behind this possibly surprising pattern is that, for some IOs, upgrades are less efficient than other bids (see Section~\ref{sec:upgrade}), situation seemingly occurring more often for high conversion probability IOs. The details about the two-slopes approximation are provided in the Appendix.

We evaluated all the agents on 1000 test episodes run in the bidding simulator, using campaigns from a period held out during the training and varying budget and target CPA. The score used to measure the agents' performance consisted in the number of conversions, penalized by the CPA exceedance coefficient (Problem~\ref{eq:simplified}). Corroborating the theoretical analysis, the oracle-upgrade outperforms the oracle-slot, although by only 3-4\% (Table~\ref{tab:oil_vs_oil}). The two-slopes approximation of the oracle-upgrade performs between the other two. Interestingly, the bids output by the oracle-slot are more effective as a target to train an auto-bidding agent with OIL. Direct regression of the oracle-upgrade's bids leads to relatively poor performance (OIL-upgrade, Table~\ref{tab:oil_vs_oil}), while imitation of the two-slopes approximation almost reaches the oracle-slot's score (OIL-upgrade-2s and OIL-slot, Table~\ref{tab:oil_vs_oil}). These results indicate that a marginally better oracle does not compensate for the difficulty of regressing bids that are different for each IO. OIL-slot's score is less than 5\% lower than the score of the oracle it imitates, indicating that it can learn a near-optimal policy even without access to the future information. As a result, OIL-slot was selected as our primary method for baseline comparisons.

We compared OIL-slot against baselines from the two main fields of research that have achieved success in real-time bidding: optimization and reinforcement learning (online and offline).

\textbf{1. Online LP.} A learning-free optimization method, implemented by \citeauthor{su2024auction}~\citep{su2024auction} based on ideas of \citeauthor{aggarwal2019autobidding}~\citep{aggarwal2019autobidding}. This algorithm uses Linear Programming to compile a table of the best bids for different values of budget and target CPA, which is used to decide how much to bid at test time.

\textbf{2. Behavior cloning (BC).} A simple algorithm mimicking bids from the AuctionNet dataset.

\textbf{3. Implicit Q-Learning (IQL).} A state-of-the-art offline RL algorithm, designed to improve over the policies used to generate the dataset using the reward signal~\citep{kostrikov2021offline}. We tested IQL with two reward functions: the score at the end of the advertisement campaign (IQL-score) and at each step (IQL-shaped).

\textbf{4. Proximal Policy Optimization (PPO).} A robust online RL algorithms~\citep{schulman2017proximal}, implemented in Stable Baselines 3~\citep{raffin2021stable}, which we evaluated in three versions: as in IQL, we considered two different reward functions (PPO-score and PPO-shaped) and we also allowed the agent to output a two-slope action (PPO-shaped-2s), as we did with OIL-upgrade-2s. This was not possible with IQL, because the bids of the policies featured in the AuctionNet are linearly proportional to the expected conversion (single slope).

Hyperparameters and experimental details are in the Appendix. For fair comparison, we allowed each algorithm to learn from 10 million environment interactions, and used the same network architecture and observation space for all experiments. 

As shown in Table~\ref{tab:results}, OIL-slot outperformed all baselines, achieving a 21.2\% and 17.83\% relative score improvement on AuctionNet-dense and AuctionNet-sparse datasets, respectively. Notably, OIL-slot balanced high conversion counts and near-complete budget utilization with low CPA. Although IQL occasionally achieved lower CPA than OIL, it consistently sacrificed almost 50\% of the conversions, resulting in lower overall score.

To validate our online training choice, we trained OIL-slot offline with the AuctionNet dataset, replacing agent bids with oracle bids. While the offline-trained agent outperformed all baselines (score: $422.43 \pm 6.12$ dense, $31.48 \pm 0.18$ sparse), its score was over 10\% lower than the online version (score: $466.36 \pm 0.43$ dense, $35.29 \pm 0.05$ sparse), confirming the advantages of online training.

\begin{figure}
    \centering
    \includegraphics[width=1\linewidth]{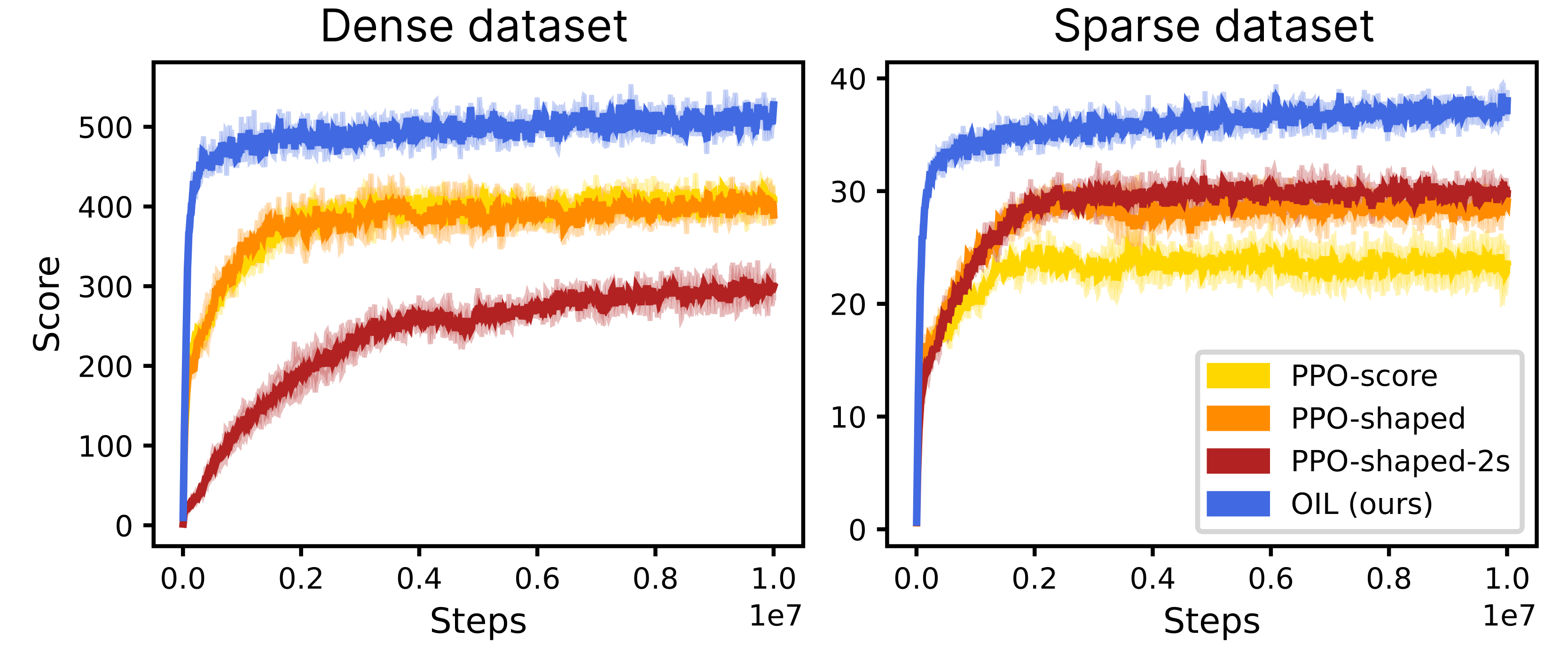}
    \caption{Learning curves PPO and OIL, AuctionNet-dense (left) and AuctionNet-sparse (right).}
    \label{fig:lc_oil_ppo}
\end{figure}

\section{Discussion, limitations and future work}

We presented OIL, an imitation learning framework to train auto-bidding agents for real-time auctions. Leveraging the structure of multi-slot impression opportunities, we developed an efficient greedy algorithm to determine near-optimal bids. Our experiments demonstrate that even a simple policy network can effectively imitate the oracle's bids, outperforming all the baselines.

Our work presents a few limitations. First, we embedded the CPA constraint into the objective function, which does not guarantee that it is always satisfied. Scenarios where this constraint might be violated include highly variable auction dynamics or cases with low availability of IOs with high conversion probability. Exploring alternative formulations which strongly impose the target CPA constraint would require significant changes in the theoretical analysis. Second, we assumed the bids of other advertisers not to be influenced by the bids of the auto-bidding agent. While this assumption is reasonable in large-scale auctions, where individual strategies have minimal collective impact, it might not hold in smaller auctions. Extending the framework to a multi-agent setting, where all advertisers' bids are jointly modeled, would provide a more realistic approach, at the cost of introducing additional complexity that is outside the scope of this study.

Although our experiments were conducted in the context of second-price auctions, OIL is well-suited for first-price auctions as well. First-price auctions also exhibit the key property of decreasing costs for successive slots within the same impression opportunity. Notably, the oracle algorithm remains unchanged when applied to first-price auctions, suggesting that OIL could achieve similar success in this setting. If large-scale first-price auction datasets become available, future studies could validate this hypothesis.

Beyond real-time bidding, the principles underlying OIL may extend to broader applications where near-optimal solutions can be efficiently derived through algorithmic approaches. Examples include budget allocation for advertisement campaigns, where the budget is distributed across multiple channels, and portfolio optimization, where investments are selected to maximize returns within predefined constraints. Future research could focus on formalizing these mappings and adapting OIL to these domains, thereby broadening its applicability and impact.

\bibliographystyle{ACM-Reference-Format}
\bibliography{bibliography}


\section*{Appendix}
\setcounter{lemma}{0}
\setcounter{theorem}{0}

\subsection*{Approximation of the MCKP}
To make Problem~\ref{eq:stochastic} tractable, we operate a few simplifications. 

If the CPA is smaller than $K$, the objective $U$ can be simplified to:
\begin{equation}
U = \mathbb{E} \left[\sum_{t,i,d} A_{t,i,d} x_{t,i,d} \right] = \sum_{t,i,d} \mathbb{E}[A_{t,i,d}]  x_{t,i,d} \simeq \sum_{t,i,d} h_d \mu_{t,i}  x_{t,i,d}.
\end{equation}
Here we have used the law of total expectations to obtain $\mathbb{E}[\beta_{t,i,d}] = h_d \mu_{t,i}$, while also assuming that $\sigma_{t,i}$ is small enough compared to $\mu_{t,i}$, so that clipping $\beta_{t,i,d}$ between 0 and 1 does not noticeably change its mean. If, instead, the CPA is larger than K, the objective function becomes:

\begin{equation}
U = \mathbb{E} \left[ K \frac{\left(\sum_{t,i,d} A_{t,i,d} x_{t,i,d}\right)^3}{\left(\sum_{t,i,d} C_{t,i,d} x_{t,i,d}\right)^2}\right] \simeq K \frac{\left(\sum_{t,i,d} h_d \mu_{t,i}  x_{t,i,d} \right)^3}{\left( \sum_{t,i,d} h_d k_{t,i,d}  x_{t,i,d} \right)^2},
\end{equation}
where we used a large number approximation (for a large number of IOs) to replace $\sum_{t,i,d} h_d \mu_{t,i}  x_{t,i,d}$ and $\sum_{t,i,d} h_d k_{t,i,d}  x_{t,i,d}$ with their expected values.

Finally, we replaced the budget constraint with a constraint in expectation, so that the oracle agent plans a strategy that, on average, does not exceed the total budget:
\begin{equation}
    \mathbb{E}\left[\sum_{t,i,d} C_{t,i,d}x_{t,i,d}\right] = \sum_{t,i,d}h_d k_{t,i,d} x_{t,i,d} \leq B
\end{equation}
We remark that, during the simulation, the budget constraint is enforced exactly, and the constraint in expectation is only used to find a solution to the problem.

With these simplifications, the optimization problem becomes:

\begin{align}
\text{maximize} \quad & U =\,\, \min\left(1, \left( K \frac{\sum_{t,i,d} h_d \mu_{t,i}  x_{t,i,d}}{\sum_{t,i,d} h_d k_{t,i,d}  x_{t,i,d}} \right)^2 \right) \left(\sum_{t,i,d} h_d \mu_{t,i}  x_{t,i,d} \right) \nonumber \\
\text{subject to} \quad & \sum_{t,i,d} h_d k_{t,i,d} x_{t,i,d} \leq B, \nonumber \\
& \sum_d x_{t,i,d} \leq 1, \quad \forall t, i, \nonumber \\
& x_{t,i,d} \in \{0, 1\},
\label{eq:simplified_app}
\end{align}

\subsection*{Two-slopes approximation (oracle-upgrade-2s)}
\begin{figure}[t]
    \centering
    \includegraphics[width=\linewidth]{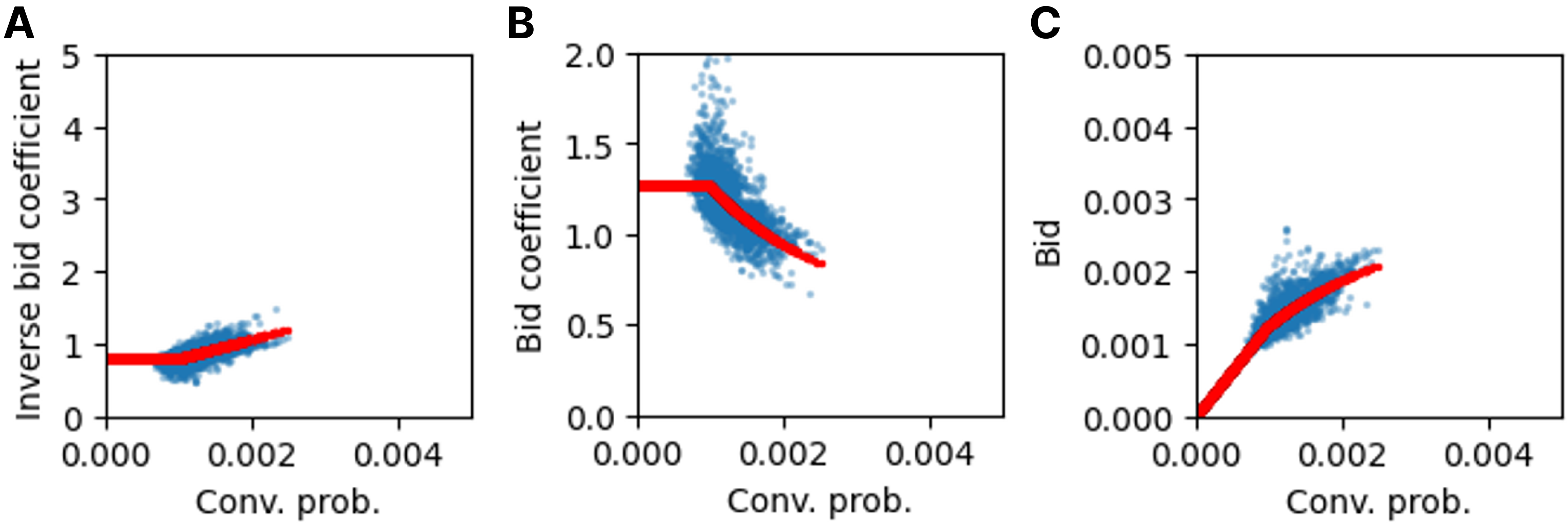}
    \caption{\textbf{A.} Two-slopes regression of the inverse bidding coefficient as a function of the conversion probability. \textbf{B.} Same as A, but for the bidding coefficients. \textbf{C} Same as A and B, but for the bids, obtained by multiplying the coefficients and the conversion probabilities.}
    \label{fig:oracle_2s}
\end{figure}
Because of the difficulties encountered by the auto-bidding agent in regressing the optimal bids output by the oracle-upgrade, we attempted to simplify the oracle's output, while preserving high performance. We observed that the ratio between the oracle's bid for an IO and the the conversion probability of that IO tend to decrease as the conversion probability gets larger. We have assumed that, if the oracle wants to win slot $d$ of IO $(t,i)$, then $b_{t,i} = \frac{k_{t,i,d} + k_{t,i,d-1}}{2}$ (i.e., the average between the price of the slot $d$ it wants to win and the price of the slot $d+1$). In fact, any bid between $k_{t,i,d}$ and $k_{t,i,d-1}$ would lead to the same outcome (winning slot $d$ of IO $(t,i)$ and paying $k_{t,i,d}$). Furthermore, the inverse of the ratio between the optimal bid and the conversion probability seems to increase linearly after a certain threshold (empirical validation, Fig.~\ref{fig:oracle_2s}A). For this reason, we decided to approximate the bids of the oracle with just three parameters: (1) The bidding coefficient for IOs with low conversion probability. The coefficient is barely small enough so that none of the inefficient slots is won and (2) the angular coefficient and (3) the intercept of the line regressing the inverse of the ratio between the oracle's bids and the conversion probabilities of the IOs that the oracle wants to win (Fig.~\ref{fig:oracle_2s}A). This simplified oracle, which we call \emph{oracle-upgrade-2s} computes the linear regression on the transformed oracle bids at every time step and finds the intercept between such line and the constant bidding coefficient for low-conversion-probability bids (Fig.~\ref{fig:oracle_2s}A). Given an IO, if the conversion probability is on the left of the crossing point, it the bidding coefficient is selected according to the low-conversion-probability line, otherwise according to the high-conversion-probability line (Fig.~\ref{fig:oracle_2s}B). The resulting bids accurately regress the average oracle-upgrade's bids for different conversion probability ranges (Fig.~\ref{fig:oracle_2s}C). The oracle-upgrade-2s outperforms the oracle-slot slightly, but it fails to provide a better training target (Table~\ref{tab:oil_vs_oil}).

\subsection*{Proofs of the theoretical results}
\sortedEfficiency

\begin{proof}
    If $i > j$, we have that
    \begin{equation}
        \eta_i > \eta_j \iff \frac{\mu}{\delta_i k_1} > \frac{\mu}{\delta_j k_1} \iff \delta_i < \delta_j,
    \end{equation}
    which is true by definition of D-SPIO.
\end{proof}

\constBidCoef

\begin{proof}
    First of all, we observe that in a second-price auction, a bid awards slot $d$ for the IO $(t,i)$ if $k_{t,i,d} \leq b_{t.i} < k_{t,i,d-1}$, i.e., the bid is larger than the cost of slot $d$, but smaller than slot $d+1$. We have used a notation abuse so that, if $d=1$, we ignore the inequality $b_{t.i} < k_{t,i,d-1}$ (any bid larger than $k_{t,i,d}$ awards the first slot). We need to prove that the bids defined by $b_{t,i} = \alpha \mu_{t,i}$ are (1) larger than the cost of all slots in $R^*$, but (2) smaller than the costs of all slots which do not belong to $R^*$.

    (1) We proceed as follows:
    \begin{align*}
        & \alpha = \max{\{1/\eta_{t,i,d} \,\,s.t.\,\, (t, i, d) \in R^*\}} \implies \\
        & \alpha \mu_{t,i} \geq \mu_{t,i} / \eta_{t,d,i} \quad \forall (t, i, d) \in R^* \implies \\
        & b_{t,i} \geq k_{t,i,d} \quad \forall (t, i, d) \in R^* \quad \text{(definition of efficiency)}
    \end{align*}
    So this means that the bidding agent bids higher than the cost of all slots in $R^*$.
    (2) We observe that all the slots that are not in $R^*$ are less efficient than the slots in $R^*$, because the slots in $R$ are ranked by efficiency and $R^*$ includes the top slots of $R$. Therefore, we have that
    \begin{align*}
        & \min{\{\eta_{t,i,d} \,\,s.t.\,\, (t, i, d) \in R^*\}} > \max{\{\eta_{t,i,d} \,\,s.t.\,\, (t, i, d) \in \overline{R^*}\}} \implies \\
        & \max{\{1/\eta_{t,i,d} \,\,s.t.\,\, (t, i, d) \in R^*\}} < \min{\{1/\eta_{t,i,d} \,\,s.t.\,\, (t, i, d) \overline{R^*} \}} \implies \\
        & \alpha \mu_{t,i} < \mu_{t,i} / \eta_{t,d,i} \quad \forall (t, i, d) \in \overline{R^*} \implies \\
        & b_{t,i} < k_{t,i,d} \quad \forall (t, i, d) \in \overline{R^*}
    \end{align*}
    where by $\overline{R^*}$ we indicated the complementary set of $R^*$, i.e., the elements of $R$ not in $R^*$. This means that the agents' bids are lower than the cost of any advertisement slot not in $R$, which concludes the proof.
\end{proof}

\dBest

\begin{proof}
    By definition of slot efficiency and upgrade efficiency we can rewrite the claim as
    \begin{align}
        \frac{\mu}{\delta_D k_1} &> \frac{\mu(h_i - h_j)}{k_1(\delta_i h_i - \delta_j h_j)} \\
        \iff \quad \frac{1}{\delta_D} &> \frac{h_i - h_j}{\delta_i h_i - \delta_j h_j}
    \end{align}
    If $h_i = h_j$ then the inequality is true. We consider the case in which $h_i \neq h_j$:
        \begin{align}
        \frac{1}{\delta_D} &> \frac{h_i - h_j}{\delta_i h_i - \delta_j h_j} \\
        \iff \quad \delta_D &< \frac{\delta_i h_i - \delta_j h_j}{h_i - h_j} \\
        \iff \quad \delta_D &< \frac{\delta_i h_i - \delta_i h_j + \delta_i h_j - \delta_j h_j}{h_i - h_j} \\
        \iff \quad \delta_D &< \delta_i + \frac{(\delta_i - \delta_j) h_j}{h_i - h_j}.
    \end{align}
    Since, by definition of D-SPIO, we have that $\delta_i \geq \delta_D$ (because $i \leq D$ by hypothesis), $\delta_i > \delta_j$ (because $i < j$ by hypothesis), $h_i > h_j$ (because $i < j$ by hypothesis) and $h_i > 0$, this inequality is always verified.
\end{proof}

\harmonic

\begin{proof}
    The cases $i=k=j$, $i=k$ and $k=j$ can be easily verified. Thus, we focus on the general case $ i < k < j$.
    Using the definition of upgrade efficiency, we have that
    \begin{align}
        \frac{1}{\tilde{\eta}_{i,j}} & = \frac{k_1 ( \delta_i h_i - \delta_j h_j )}{\mu ( h_i - h_j)} \\
        & = \frac{k_1(\delta_i h_i - \delta_k h_k) + k_1(\delta_k h_k - \delta_J h_j)}{\mu (h_i - h_j)} \\
        & = \frac{h_i - h_k}{h_i - h_j} \frac{k_1 (\delta_i h_i - \delta_k h_k)}{\mu ( h_i - h_k)} + \frac{h_k - h_j}{h_i - h_j} \frac{k_1 (\delta_h h_k - \delta_h h_j)}{\mu(h_k - h_j)} \\
        & = \frac{h_i - h_k}{h_i - h_j} \frac{1}{\tilde{\eta}_{i,k}} + \frac{h_k - h_j}{h_i - h_j} \frac{1}{\tilde{\eta}_{k,j}}.
    \end{align}
    Since
    \begin{equation}
        \frac{h_i - h_k}{h_i - h_j} + \frac{h_k - h_j}{h_i - h_j} = 1,
    \end{equation}
    the first claim is proven. For the second claim, we can simply observe that
    \begin{align}
        & \tilde{\eta}_{i,k} \geq \tilde{\eta}_{k,j} \implies \frac{1}{\tilde{\eta}_{i,k}} \leq \frac{1}{\tilde{\eta}_{k,j}} \implies \\
        &\beta \frac{1}{\tilde{\eta}_{i,k}} + (1-\beta) \frac{1}{\tilde{\eta}_{k,j}} \leq \frac{1}{\tilde{\eta}_{k,j}} \implies \\
        & \frac{1}{\tilde{\eta}_{i,j}} \leq \frac{1}{\tilde{\eta}_{k,j}} \implies \tilde{\eta}_{i,j} \geq \tilde{\eta}_{k,j}.
    \end{align}
    As the argument works also swapping $\tilde{\eta}_{i,k}$ and $\tilde{\eta}_{k,j}$, the second claim is proven.
\end{proof}

\optimalityGap

\begin{proof}
    For each slot $d$ of the IO $(t, i)$, we define the upgrade conversion probability $\Delta m_{t,i,d} = \mu_{t,i} (h_d - h_{d+1})$ and the upgrade cost $\Delta c_{t,i,d} = k_{t,i,d} h_d - k_{t,i,d+1}h_{d+1}$ (we consider $h_{D+1} = k_{t,i,D+1} = 0$ for all impression $(t,i)$, because there are only $D$ slots. We also define the upgrade efficiency $\tilde{\eta}_{t,i,d} = \Delta m_{t,i,d} / \Delta c_{t,i,d}$). For ease of notation, we have assumed (without loss of generality) that the upgrades within slots of the same IO are sorted by efficiency. If this is not the case, it is sufficient to merge upgrades as explained in Section~\ref{sec:upgrade} to restore the order. By construction, the solution set $R^*$ provided by the oracle-upgrade algorithm guarantees that $\tilde{\eta}_{t,i,d} \geq \tilde{\eta}_{t',i',d'} \forall (t,i,d) \in R^*, (t',i',d') \notin R^*$. Therefore, if $\sum_{t,i,d}h_d k_{t,i}x_{t,i,d} = B$ the solution $x_{t,i,d} = \mathbbm{1}_{R^*}(t,i,d)$ is optimal, because replacing any slot upgrade in $R^*$ with one outside of $R^*$ would cause a decrease in efficiency and therefore lower expected total conversions for the same (or lower) budget expenditure. It follows that any improvement over the solution provided by the oracle-upgrade algorithm must have a higher budget consumption.

    Let us consider the optimal solution $x_{t',i',d'} = \mathbbm{1}_{S}(t',i',d')$ which collects the advertisement slots in $S$. We know that the budget consumption of the optimal solution is larger or equal than the budget consumption of the oracle solution, but still within the budget constraint:
    \begin{equation}
        \sum_{(t,i,d)\in R^*}h_d k_{t,i}x_{t,i,d} \leq \sum_{(t',i',d')\in S}h_d k_{t',i'}x_{'t,i',d'} \leq B.
        \label{eq:budget_ineq}
    \end{equation}
    Furthermore, the efficiency of the optimal solution cannot be higher than the one of the upgrade-oracle solution, because all the slot upgrades not in $R^*$ that could be used to replace slot upgrades in $R^*$ are less efficient than those in $R^*$, and cannot improve the overall efficiency while also increasing the budget consumption. Therefore we have that 
    \begin{equation}
        \eta_S \leq \eta_{R^*},
        \label{eq:eff_ineq}
    \end{equation}
    where
    \begin{equation}
        \eta_S = \frac{\sum_{(t',i',d')\in S} h_{d'} \mu_{t',i'}x_{t',i',d'}}{\sum_{(t',i',d')\in S}h_{d'} k_{t',i'}x_{t',i',d'}}
    \end{equation}
    and
    \begin{equation}
         \eta_{R^*} = \frac{\sum_{(t,i,d)\in R^*}h_d \mu_{t,i}x_{t,i,d}}{\sum_{(t,i,d)\in R^*}h_d k_{t,i}x_{t,i,d}}.
    \end{equation}

    By combining Eq. \ref{eq:budget_ineq} and \ref{eq:eff_ineq}, we obtain the following estimate of the optimality gap:
    \begin{align}
        & U_{\max} - U = \sum_{(t',i',d')\in S} h_{d'} \mu_{t',i'}x_{t',i',d'} - \sum_{(t,i,d)\in R^*}h_d \mu_{t,i}x_{t,i,d} \\
        & = \eta_S \sum_{(t',i',d')\in S}h_d k_{t',i'}x_{'t,i',d'} - \eta_{R^*} \sum_{(t,i,d)\in R^*}h_d k_{t,i}x_{t,i,d} \\
        &\leq \eta_{R^*} \left(B - \sum_{(t,i,d)\in R^*}h_d k_{t,i}x_{t,i,d} \right) = \eta_{R^* }\Delta B,
    \end{align}
     where $\Delta B = B - \sum_{t,i,d}h_d k_{t,i}x_{t,i,d}$ is the expected spare budget at the end of the advertisement campaign. This concludes the proof.
\end{proof}

\subsection*{Notation}
Table~\ref{tab:notation} summarizes the main symbols recurring in this paper.

\begin{table}[]
\centering
\caption{Notation used in the paper.}
\label{tab:notation}
\resizebox{1\linewidth}{!}{
    \begin{tabular}{cl}
    \toprule
    \textbf{Symbol} & \textbf{Explanation} \\
    \midrule
    \multicolumn{2}{c}{\textit{General}} \\
    $B$ & Total campaign budget \\
    $K$ & Target cost-per-acquisition (CPA) \\
    $\mu$ & Mean conversion probability \\
    $\sigma$ & Standard deviation of conversion probability \\
    $N$ & Number of impression opportunities (IOs) \\
    $D$ & Number of advertisement slots per IO \\
    $t$ & Time step in the campaign \\
    $i$ & Index of an impression opportunity (IO) \\
    $d$ & Index of an advertisement slot \\
    $h_d$ & Exposure probability for slot $d$ \\
    $c_t$ & Total cost incurred at time $t$ \\
    $A_{t,i,d}$ & Binary variable for acquisition event (1 = conversion) \\
    $C_{t,i,d}$ & Cost for winning slot $d$ of IO $i$ at time $t$ \\
    $x_{t,i,d}$ & Binary variable indicating if slot $d$ is won at $t$ \\
    $R^*$ & Optimal set of slots selected by the oracle \\
    \\
    \multicolumn{2}{c}{\textit{Efficiency}} \\
    $\eta_{t,i,d}$ & Efficiency of slot $d$ of IO $i$ at time $t$ (e.g., $p_d / c_d$) \\
    $\eta_{i,j}$ & Efficiency of upgrading from slot $j$ to slot $i$ \\
    $\Delta c_{t,i,d}$ & Cost difference for upgrading slots \\
    $\Delta m_{t,i,d}$ & Conversion probability difference for upgrading slots \\
    \\
    \multicolumn{2}{c}{\textit{Performance Metrics}} \\
    $U$ & Objective of the MCKP \\
    CPA & Cost-per-acquisition: $\text{CPA} = \frac{\sum C_{t,i,d} x_{t,i,d}}{\sum A_{t,i,d} x_{t,i,d}}$ \\
    $\text{Score}$ & Performance metric: total acquisitions scaled by CPA penalty \\
    \\
    \bottomrule
    \end{tabular}
    }
\end{table}

\subsection*{Observation features}
Table~\ref{tab:observation_features} lists the features extracted from the real-time data provided to all the agents trained for this work.

\begin{table}[]
\centering
\caption{Observation Features Used by the Auto-Bidding Agent. LWC = least winning cost, IO = impression opportunity.}
\label{tab:observation_features}
    \resizebox{0.9\linewidth}{!}{

        \begin{tabular}{cl}
        \toprule
        \textbf{\#} & \textbf{Feature Explanation} \\
        \midrule
        1  & Time left in the campaign \\
        2  & Budget remaining for the campaign \\
        3  & Total campaign budget \\
        4  & Current cost-per-acquisition (CPA) \\
        5  & Category of the campaign \\
        6  & Mean of historical bids \\
        7  & Mean of bids from the last step \\
        8  & Mean of bids from the last three steps \\
        9  & Mean least winning cost (LWC) \\
        10 & Mean LWC from the last step \\
        11 & Mean LWC from the last three steps \\
        12 & 10th percentile of LWC \\
        13 & 10th percentile of LWC from the last step \\
        14 & 10th percentile of LWC from the last three steps \\
        15 & 1st percentile of LWC \\
        16 & 1st percentile of LWC from the last step \\
        17 & 1st percentile of LWC from the last three steps \\
        18 & Mean conversion probability (pvalue) \\
        19 & Mean historical conversion rate \\
        20 & Mean historical bid success rate \\
        21 & Mean pvalue from the last step \\
        22 & Mean pvalue from the last three steps \\
        23 & Mean conversion rate from the last step \\
        24 & Mean conversion rate from the last three steps \\
        25 & Bid success rate from the last step \\
        26 & Mean bid success rate from the last three steps \\
        27 & Mean position of successful historical bids \\
        28 & Mean position of successful bids from the last step \\
        29 & Mean position of successful bids from the last three steps \\
        30 & Mean historical cost \\
        31 & Mean cost from the last step \\
        32 & Mean cost from the last three steps \\
        33 & Mean historical cost for slot 1 \\
        34 & Mean cost for slot 1 from the last step \\
        35 & Mean cost for slot 1 from the last three steps \\
        36 & Mean historical cost for slot 2 \\
        37 & Mean cost for slot 2 from the last step \\
        38 & Mean cost for slot 2 from the last three steps \\
        39 & Mean historical cost for slot 3 \\
        40 & Mean cost for slot 3 from the last step \\
        41 & Mean cost for slot 3 from the last three steps \\
        42 & Mean historical bid over LWC ratio \\
        43 & Bid over LWC ratio from the last step \\
        44 & Bid over LWC ratio from the last three steps \\
        45 & Historical ratio of pvalue over LWC \\
        46 & Pvalue over LWC ratio from the last step \\
        47 & Pvalue over LWC ratio from the last three steps \\
        48 & 90th percentile of pvalue over LWC \\
        49 & 90th percentile of pvalue over LWC from the last step \\
        50 & 90th percentile of pvalue over LWC from the last three steps \\
        51 & 99th percentile of pvalue over LWC \\
        52 & 99th percentile of pvalue over LWC from the last step \\
        53 & 99th percentile of pvalue over LWC from the last three steps \\
        54 & Current mean pvalue \\
        55 & 90th percentile of current pvalues \\
        56 & 99th percentile of current pvalues \\
        57 & Number of current IOs \\
        58 & Number of IOs from the last step \\
        59 & Number of IOs from the last three steps \\
        60 & Total number of IOs \\
        \bottomrule
        \end{tabular}
    }
\end{table}

\subsection*{Hyperparameters}
The hyperparameters of OIL and of the baselines are listed in Table~\ref{tab:hyperparameters}.
\begin{table}[]
    \centering
    \caption{Hyperparameters of the algorithms featured in this work.}
    \label{tab:hyperparameters}
    \resizebox{0.79\linewidth}{!}{
        \begin{tabular}{lll}
            \toprule
            \textbf{Algorithm} &\textbf{Hyperparameter} & \textbf{Value} \\
            \midrule
            OIL  & Learning rate (start) & $10^{-3}$\\
                 & Learning rate (end) & 0 \\
                 & Learning rate schedule & Linear \\
                 & Batch size & 512 \\
                 & Rollout steps & 128\\
                 & Num epochs & 10\\
                 & Entropy coefficient & $3\times 10^{-6}$\\
                 & Max gradient norm & 0.7\\
                 & Policy hidden layers & [256, 256, 256] \\
                 & Activation & ReLU \\
                 & Initial action std & 1 \\
            \midrule
             PPO & Learning rate (start) & $10^{-4}$\\
                 & Learning rate (end) & 0 \\
                 & Learning rate schedule & Linear \\
                 & Batch size & 512 \\
                 & Rollout steps & 128\\
                 & Num epochs & 10\\
                 & Discount factor $\gamma$ & 0.99 \\
                 & Entropy coefficient & $3\times 10^{-6}$\\
                 & Value function coefficient & 0.5\\
                 & GAE $\lambda$ & 0.9\\
                 & Clip parameter & 0.3\\
                 & Max gradient norm & 0.7\\
                 & Policy hidden layers & [256, 256, 256] \\
                 & Critic hidden layers & [256, 256, 256] \\
                 & Activation & ReLU \\
                 & Initial action std & 1 \\
            \midrule
            Online LP & Cost interval & 10 \\
                      & Subsample fraction & $1 / 150$ \\
            \midrule
            BC   & Learning rate (start) & $10^{-4}$\\
                 & Learning rate (end) & $10^{-4}$ \\
                 & Learning rate schedule & Constant \\
                 & Batch size & 100 \\
                 & Policy hidden layers & [256, 256, 256] \\
                 & Activation & ReLU \\
            \midrule
            IQL  & Learning rate (start) & $10^{-4}$\\
                 & Learning rate (end) & $10^{-4}$ \\
                 & Learning rate schedule & Constant \\
                 & Batch size & 100 \\
                 & Policy hidden layers & [256, 256, 256] \\
                 & V hidden layers & [256, 256, 256] \\
                 & Q hidden layers & [256, 256, 256] \\
                 & Activation & ReLU \\
                 & Discount factor $\gamma$ & 0.99 \\
                 & Soft update coef. $\tau$ & 0.01 \\
                 & Expectile & 0.7 \\
                 & Temperature & 3 \\
            \bottomrule
        \end{tabular}
    }
\end{table}

\end{document}